\pdfoutput=1
\documentclass{article}

\usepackage[final]{corl_2026} 
\makeatletter
\renewcommand{\@noticestring}{SPAIS Workshop at the \@conferenceordinal\/ Conference on Robot Learning (CoRL \@conferenceyear), \@conferencelocation.}
\makeatother
\usepackage{amsmath,amssymb}
\usepackage{graphicx}
\usepackage{booktabs}
\usepackage{enumitem}
\usepackage{multirow}
\usepackage[font=small,skip=3pt]{caption}
\usepackage[font=small,skip=1pt]{subcaption}
\newcommand{\num}[1]{#1}

\usepackage{colortbl}
\usepackage{tikz}
\usetikzlibrary{arrows.meta}

\title{Right Goal, Wrong Body: Re-grounding Frozen VLAs under Actuation Shift}

\author{
  Umer Siddique, Peilang Li, Yongcan Cao\\
  Department of Electrical and Computer Engineering\\
  University of Texas at San Antonio, United States\\
  \texttt{muhammadumer.siddique@my.utsa.edu}
}

\begin{document}
\maketitle


\begin{abstract}
Vision-language-action (VLA) models are capable of producing desired robotic behaviors in ideal settings. However, they often fail when the robot's actuation changes at deployment. For example, remounting, mechanical wear, or command latency changes can lead to failed tasks and unsafe behaviors. To address the inevitable \emph{actuation shift}, we propose ReGround, an inference-time method that keeps the VLA frozen and re-grounds its intended motion. ReGround learns the deployed robot's dynamics from task-agnostic motor babbling, and then tracks the policy's intended motion while staying close to the policy's own plan. ReGround activates only upon detecting a shift, and needs no rewards, demonstrations, or retraining. We show via two VLAs, eight LIBERO tasks, and five shift settings that ReGround can restore near-oracle success, such as raising $\pi_{0.5}$'s success from 0.16 to 0.90 under 400\,ms latency, while remaining safer than a linear-calibration baseline. Project website and open-source code: \href{https://umersiddique94.github.io/ReGround/}{\texttt{umersiddique94.github.io/ReGround}}.
\end{abstract}


\section{Introduction}
\label{sec:intro}
Recently, vision-language-action (VLA) models have achieved strong performance across diverse robotic manipulation tasks by leveraging large-scale robot datasets and vision-language
representations to map observations and language instructions directly to actions~\citep{brohan2023rt1,brohan2023rt2,oxe2024,octo2024,kim2024openvla,black2024pi0,intelligence2025pi05,shukor2025smolvla}. 
However, a VLA is typically trained assuming a particular relationship between its action commands and the robot's resulting motion. 
Changes in the action coordinate frame, controller gains, or command latency can alter this relationship, causing the same action predicted by the VLA to produce a different physical motion. 
We refer to this mismatch as \emph{actuation shift}, where the task and observations remain unchanged, but the mapping from the VLA's action commands to the robot's physical motion has changed. 
A natural solution is to retrain or fine-tune the policy on the deployed robot, but this requires additional task demonstrations, computation, and access to model weights~\citep{julian2020never,tobin2017domain,peng2018sim,zhao2020sim}. 
This motivates our central question: \emph{At deployment, how can a frozen VLA remain effective and safe when the robot's actuation has changed?}

Actuation shift is not only a performance issue but also a safety concern~\citep{liu2024safety}. A VLA may still be able to identify the correct task and target object while producing actions that lead to unintended robot motion, resulting in overshoot, unintended contact, object displacement, or other unsafe behavior.  
Existing work on VLA robustness has primarily studied changes in visual observations~\citep{fei2025liberoplus}, while adaptation to changes in robot actuation generally relies on additional fault data, demonstrations, or policy updates~\citep{jo2026uncovering}. 
Training-time approaches, such as rapid motor adaptation and meta-learning, improve robustness by exposing policies to variations during training or adapting them after deployment~\citep{kumar2021rma,nagabandi2019learning}. 
More recently, test-time methods have also shown that frozen policies can be steered without parameter updates by using reward, value, or verifier signals to guide action selection~\citep{nakamoto2024steering,kwok2025robomonkey,dai2025rover}. Closest to our work is the
inverse-dynamics transfer, which maps intended motion to actions under changed execution dynamics~\citep{christiano2016transfer}. However, these methods were not designed for frozen VLAs that output action chunks, nor do they address actuation shift.

We address this gap with \textbf{ReGround}, an inference-time method for adapting frozen VLAs under actuation shift. 
Our key insight is that actuation shift changes how the robot executes the VLA's actions, while the frozen VLA still knows what motion is needed to complete the task. 
ReGround therefore recovers the motion intended by the VLA under the training-body dynamics and computes new actions that reproduce this motion on the deployed robot.
To achieve this, a model of the training body rolls out the VLA's action chunk to obtain a reference trajectory. A deployed-body model learned from task-agnostic motor babbling is then used to determine how the robot should act to follow this trajectory under the changed dynamics. A trust region then keeps the corrected actions close to the VLA's original plan, while a residual monitor activates ReGround only when an actuation shift is detected.
On $\pi_{0.5}$ across eight LIBERO tasks and five shift settings, ReGround restores 0.90--0.98 success, including 0.16 to 0.90 at 400\,ms latency. For both VLAs, ReGround reduces unsafe episodes relative to linear calibration. 

\section{Preliminaries}
\label{sec:prelim}

\noindent\textbf{Vision Language Action Models.}
A VLA is typically trained via imitation learning on diverse robot demonstration datasets $\mathcal{D}$, by maximizing $\mathbb{E}_{\mathcal{D}}[\log \pi_\theta (a_{t:t+T-1} \mid o_t, \ell)]$ that maps observation $o_t$ and an instruction $\ell$ to a chunk of $T{=}50$ actions, of which the first $n{=}10$ are executed before re-querying~\citep{black2024pi0,shukor2025smolvla}. Each $a=(u,g)$ is a 6-D EEF delta for an operational-space controller plus a gripper command. We define robot state as the \emph{robot} EEF pose in 6-D rotation form, gripper state, joint positions, and velocities (25-dimensional in total), which evolves under dynamics $P(x'\mid x,a)$~\citep{Puterman94,sutton1998reinforcement} and since $x$ contains only the \emph{robot}, every model we learn is task-agnostic. To denote the robot training and deployed bodies, we use $P_{\mathrm{nom}}$ and $P_{\mathrm{dep}}$, respectively.

\noindent\textbf{Inference-time alignment and policy shaping.}
Inference-time alignment methods shape a policy at test time without updating its weights. Formally, these methods solve $\max_{\pi'}\mathbb{E}_{\pi'}[r]-\beta\,\mathrm{KL}(\pi'\,\Vert{}\,\pi_{\mathrm{ref}})$, whose optimum tilts the reference by the reward,
\begin{equation}
\pi'(a\mid s)\ \propto\ \pi_{\mathrm{ref}}(a\mid s)\,\exp\!\big(r(s,a)/\beta\big).
\label{eq:kl}
\end{equation}
These methods realize $\pi'$ by sampling or controlled decoding, without updating the weights of $\pi_{\mathrm{ref}}$~\citep{beirami2024theoretical,shi2024decoding,mujtaba2026aligning,siddique2026fairness}. The KL term acts as a trust region, similar to conservative policy updates in RL~\citep{kakade2002approximately,SchulmanWolskiDhariwalRadfordKlimov17}. However, these methods require a reward model, and no task reward is available when a robot's actuation shifts at deployment. In this paper, we formulate inference-time alignment as re-grounding base policy actions via two key substitutions. First, the reward is \emph{tracking the reference policy's intended motion}. Second, the KL divergence is replaced with a quadratic trust-region term around the reference action, thereby removing the need for an external reward model.

\section{ReGround: Re-grounding a Frozen Policy's Intent} \label{sec:method}
\begin{figure}[t]
\centering
\newsavebox{\boxA}\newsavebox{\boxB}
\newlength{\figH}\setlength{\figH}{0.245\linewidth}
\sbox{\boxA}{\includegraphics[height=\figH]{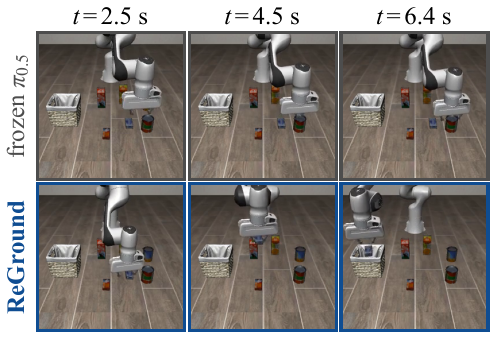}}
\sbox{\boxB}{\includegraphics[height=\figH]{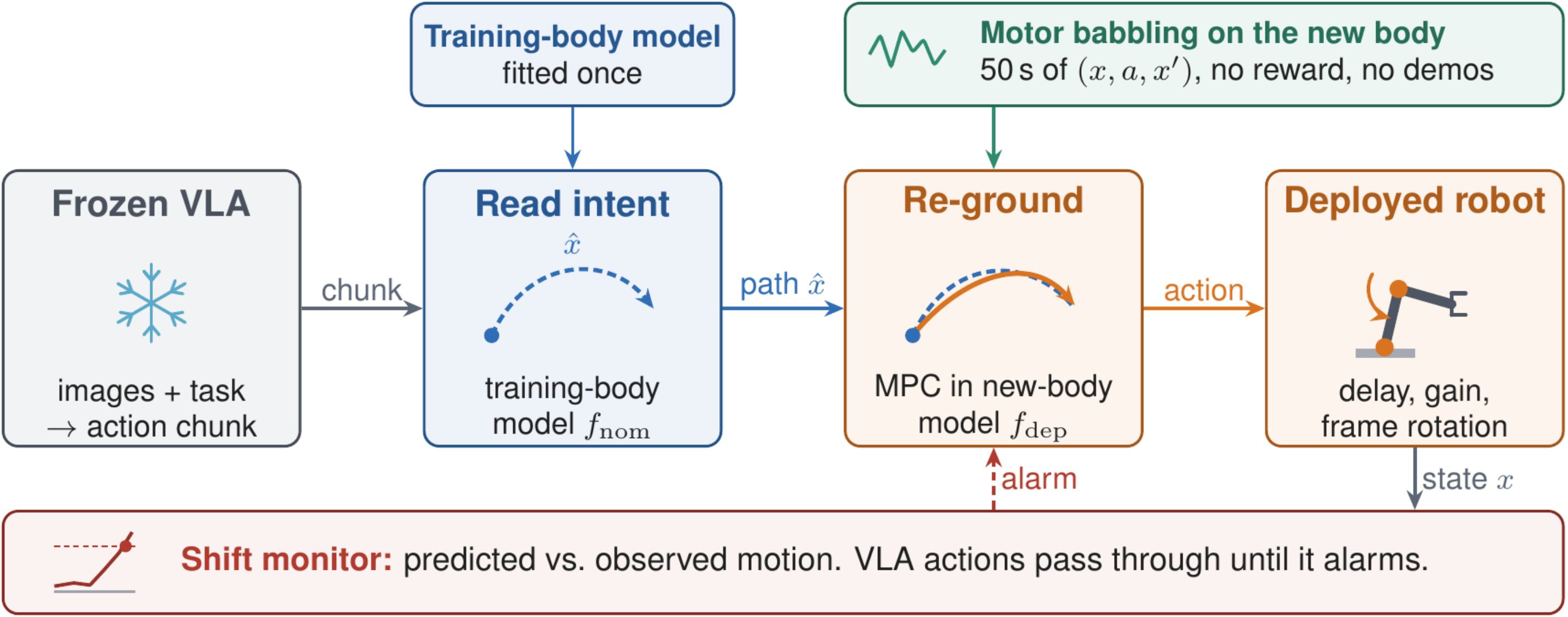}}
\begin{subfigure}[b]{\wd\boxA}
\usebox{\boxA}
\caption{Latency 400\,ms (paired episode).}
\label{fig:teaser}
\end{subfigure}\hfill
\begin{subfigure}[b]{\wd\boxB}
\usebox{\boxB}
\caption{ReGround.}
\label{fig:method-diagram}
\end{subfigure}
\caption{Right goal, wrong body. (a) Frozen $\pi_{0.5}$ with and without ReGround under 400\,ms actuation delay. (b) ReGround reads the intended path with the training-body model $f_{\mathrm{nom}}$ (blue), re-grounds it with a trust-region MPC in the deployed-body model $f_{\mathrm{dep}}$ (orange), and a CUSUM monitor enables it only after a shift.}
\label{fig:method}
\end{figure}

We now present our method, which aligns the intent of a frozen VLA at test time without updating the parameters. Specifically, our objective is to construct an inference-time policy $\pi'$ whose trajectory on the deployed body matches the frozen policy's trajectory under nominal dynamics, while making the smallest necessary action change without updating the parameters of the base policy:
\begin{equation}
\pi'=\arg\min_{\pi'}\mathbb{E}\left[|\tau_{\mathrm{dep}}(\pi')\ominus\tau_{\mathrm{nom}}(\pi)|^2\right]+\lambda D(\pi'|\pi),
\label{eq:objective}
\end{equation}
where $\tau_{\mathrm{nom}}(\pi)$ is the trajectory the frozen VLA follows in the training body, $\pi$ is the base VLA policy, and $\pi'$ is the shaped policy. This is similar to inference-time alignment, with trajectory tracking replacing the reward. Although solving~\eqref{eq:objective} seems easy and conceptually general, it presents three key challenges: (i) how to infer intent without a task reward (\textit{i.e.,} defining $\tau_{\mathrm{nom}}(\pi)$ at test time); (ii) how to identify unknown deployment dynamics $P_{\mathrm{dep}}$ without task demonstrations; and (iii) ensuring that inference-time alignment does no harm when nothing has shifted (preserving nominal behavior). We address each challenge below (Fig.~\ref{fig:method}).

\textbf{Intent read-out.} To infer intent without an external task signal, we roll out the policy chunk $u_{0:T-1}$ through a forward model of the training body, fitted once on nominal babbling:
$\hat x_0=x_t$, $\hat x_{k+1}=\hat x_k\oplus f_{\mathrm{nom}}(\hat x_k,h_k,u_k), \quad k < T$, where $\oplus$ composes pose increments, and $h_k$ is the recent action history. The resulting $T$-step path $\hat x_{1:T}$ represents the motion the policy intended to produce, even though only the first $n{=}10$ actions are executed before replanning.

\textbf{Deployed-body identification.} We collect a rollout of state-action-next-state tuples $(x,a,x')$ by smooth Ornstein--Uhlenbeck motor babbling inside a workspace box, optionally mixing in VLA-driven transitions. We fit an ensemble forward model $f_{\mathrm{dep}}$ with joint velocities and action history, and identify actuation delay $\hat d$ and a linear action map $M$ for the control anchor. This requires no task labels, rewards, or demonstrations. 
The forward model is trained to predict rotation-vector pose increments, while $\hat d$ and $M$ calibrate the action anchor.

\textbf{Trust-region tracking.} Finally, to make the deployed body follow the intended path while changing the policy's actions, we approximate the problem~\eqref{eq:objective} with a receding-horizon controller. At each control step $k$, a short-horizon model minimizes
\begin{equation}
\sum_{j=1}^{H}\gamma^j\big\|W(\tilde x_j\ominus\hat x_{k+j})\big\|^2+\lambda\sum_{j=0}^{H-1}\|u_j-\bar u_j\|^2,
\quad \tilde x_0=x_k,\ \ \tilde x_{j+1}=\tilde x_j\oplus f_{\mathrm{dep}}(\tilde x_j,\tilde h_j,u_j),
\label{eq:mpc}
\end{equation} 
where $\gamma$ is the discount factor, $H$ is the planning horizon, $W$ normalizes position and orientation, and $\tilde h_j$ is the action history along the imagined trajectory. In~\eqref{eq:mpc}, the first term is the tracking reward that replaces $r$ in~\eqref{eq:kl}. This penalizes deviation from the intended path $\hat x$ under the deployed-body model. The second term is the trust region that replaces the KL divergence, as it keeps the actions close to the anchor $\bar u$, which is the policy's own chunk re-timed by $\hat d$ and re-mapped by $M$ (or the one-step inverse-dynamics action). Thus, the weight $\lambda$ plays the role of $\beta$. In settings when $\lambda{=}0$, ReGround reduces to pure model-based tracking. As $\lambda\to\infty$, it executes the re-grounded anchor. We solve~\eqref{eq:mpc} with batched CEM, execute the first action $u^*_0$ together with the VLA's gripper command, and re-solve at the next step. In practice, on our shifts, the $\lambda$ selected on validation keeps the solution at the anchor, and $H{=}1$ matches longer horizons in success. The gains therefore come from deployed-body adaptation that preserves the policy's re-grounded plan. 

\textbf{Shift monitor.} To address whether ReGround does any harm when nothing is changed, we propose that ReGround stays inactive until a shift is detected. A one-sided CUSUM test accumulates the normalized one-step residual $e_t$ of the training-body model, $S_t=\max(0,S_{t-1}+e_t-\mu-\kappa)$, with offset $\mu$ and slack $\kappa$, and raises an alarm when $S_t$ exceeds a threshold calibrated on nominal validation episodes. Before the alarm, VLA's actions are executed unchanged. ReGround is the identity on an unshifted body. After that, its re-grounding is enabled.

\begin{figure}[t]
  \centering
  \includegraphics[width=0.85\linewidth]{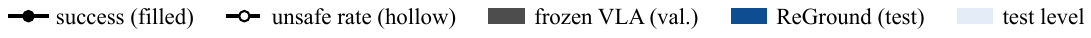}\\[1pt]
  \includegraphics[width=\linewidth]{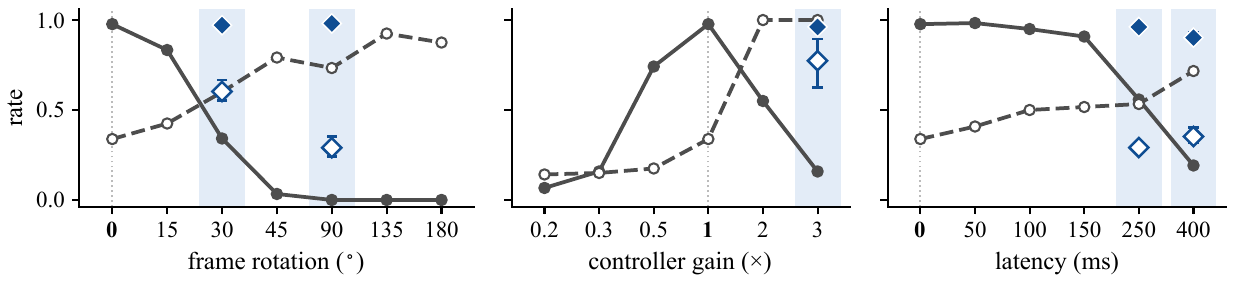}
  \caption{Demonstration of actuation-shift for frozen $\pi_{0.5}$ across 8 Libero tasks. Success is shown in filled and solid while unsafe rate is shown in hollow and dashed.}
  \label{fig:ladder}
\end{figure}

\section{Experimental Results}

We evaluate frozen $\pi_{0.5}$ and SmolVLA on eight LIBERO tasks (MuJoCo, robosuite control at 20\,Hz)~\citep{liu2023libero,intelligence2025pi05,shukor2025smolvla}. We alter the actuation while keeping visual observations unchanged, covering action-frame rotation, controller gain, latency, and a composite shift. We compare ReGround against the base VLA (Base), linear calibration (Linear), one-step inverse-dynamics transfer (Intent)~\citep{christiano2016transfer}, and the analytic inverse (Oracle). Hyperparameters are selected on validation states 0-14 and reported on disjoint test states 15-49, with 3-5 adaptation seeds and paired VLA noise. Success and unsafe rates are task-seed IQMs with stratified bootstrap intervals~\citep{agarwal2021deep}; an episode is unsafe if it involves non-target contact, object displacement or falls, or a speed-envelope violation. We ask: (A) Does actuation shift break frozen VLAs and induce unsafe behavior? (B) Does ReGround restore task success? (C) Is ReGround safe, efficient, and lightweight to deploy?

\textbf{Question A} 
Fig. \ref{fig:ladder} shows that validation success falls from 0.99 under nominal conditions to 0.19 at 400\,ms latency, 0.00 at 90$^\circ$ rotation, 0.16 at gain $\times3$, and 0.00 on the composite shift across all shift families. Concurrently, the unsafe rate rises from 0.09 nominally to 0.54 and 0.82 under severe shifts, confirming that actuation shift severely breaks frozen policies and compromises physical safety. This highlights that VLAs degrade quickly as the shift grows, motivating methods that restore both performance and safety.

\begin{table}[h]
\centering
\caption{Success and unsafe-episode rate during testing phase in terms of IQM over 8 tasks.}
\label{tab:main}
\small
\setlength{\tabcolsep}{0pt}
\renewcommand{\arraystretch}{0.85}
\begin{tabular*}{\linewidth}{@{\extracolsep{\fill}} l ccc>{\columncolor{blue!7}}cc c ccc>{\columncolor{blue!7}}c @{}}
\toprule
 & \multicolumn{5}{c}{Success $\uparrow$} & & \multicolumn{4}{c}{Unsafe-episode rate $\downarrow$}\\
\cmidrule(lr){2-6}\cmidrule(l){8-11}
Shift & Base & Linear & Intent & \textbf{ReGround} & Oracle & & Base & Linear & Intent & \textbf{ReGround}\\
\midrule
\multicolumn{11}{@{}l}{\cellcolor{white}\textit{$\pi_{0.5}$}}\\
Latency 250\,ms      & 0.61 & 0.49 & 0.80 & \textbf{0.96} & 0.97 & & 0.68 & 1.00 & \textbf{0.17} & 0.29\\
Latency 400\,ms      & 0.16 & 0.07 & 0.16 & \textbf{0.90} & 0.91 & & 0.86 & 1.00 & 0.45 & \textbf{0.35}\\
Composite  & 0.00 & 0.86 & 0.67 & \textbf{0.97} & 0.92 & & 0.78 & 0.23 & \textbf{0.15} & 0.17\\
Rotation 90$^\circ$  & 0.00 & \textbf{0.98} & \textbf{0.98} & \textbf{0.98} & 0.98 & & 0.80 & 0.76 & 0.36 & \textbf{0.29}\\
Gain $\times$3       & 0.15 & 0.94 & \textbf{0.98} & 0.96 & 0.98 & & 1.00 & 1.00 & 0.80 & \textbf{0.77}\\
\midrule
\multicolumn{11}{@{}l}{\cellcolor{white}\textit{SmolVLA} (not retuned)}\\
Latency 250\,ms      & 0.19 & 0.00 & 0.00 & \textbf{0.93} & -- & & 0.54 & 0.76 & \textbf{0.03} & 0.05\\
Composite  & 0.00 & 0.54 & 0.49 & \textbf{0.78} & -- & & 0.29 & 0.25 & \textbf{0.08} & 0.10\\
Rotation 30$^\circ$  & 0.01 & \textbf{0.91} & 0.90 & 0.89 & -- & & 0.48 & 0.12 & \textbf{0.06} & 0.10\\
Gain $\times$2       & 0.14 & 0.79 & \textbf{0.88} & 0.78 & -- & & 0.75 & 0.24 & \textbf{0.12} & 0.15\\
\bottomrule
\end{tabular*}
\end{table}

\textbf{Question B} We perform experiments and report average IQM in Table~\ref{tab:main}. Our results show that on $\pi_{0.5}$, across all five cells, ReGround recovers \num{0.91-1.00} of lost success, achieving overall success of \num{0.90-0.98} and matching the oracle. Under timing shifts, where no static map recovers success, ReGround outperforms the best baseline by \num{+0.16} at 250\,ms and by \num{+0.74} at 400\,ms, and by \num{+0.11} on the composite shift, exceeding even the analytic inverse (\num{0.97} vs.\ \num{0.92}). On rotation $90^\circ$ and gain $\times3$, a static linear map already restores success because these shifts are static re-mappings of the action; there ReGround matches its success while lowering the unsafe rate from 0.76 to 0.29 (rotation) and from 1.00 to 0.77 (gain), though Intent is safer in several cells, mostly at lower success. Per-task results are given in Appendix~\ref{app:extra}.

\textbf{Question C} 
To answer this question, we perform an experiment and show that ReGround does no harm and is safe to deploy. Our results show the success rate of 0.98 vs.\ 0.97 for the frozen VLA, at the same unsafe rate (0.40--0.41), whether it is always on or switched on by the monitor. The monitor catches every shifted episode after the shift starts, and false-alarms on 2.1\% of unshifted ones (Fig.~\ref{fig:effects}d, Appendix~\ref{app:extra}). At 250\,ms latency, 1k babbling transitions (50\,s of robot time) already give 0.98 success, where intent map needs 10k (Fig.~\ref{fig:effects}c, Appendix~\ref{app:extra}).

\section{Conclusion}
\label{sec:conclusion}
In this paper, we studied frozen robot foundation models under actuation shift: right goal, wrong body. The key to our method is that it reads the policy's intent out with a training-body forward model, re-grounds it with a deployed-body model learned from motor babbling and a trust region around the policy's actions, and switches on only when a monitor detects the shift. Through extensive experiments on two VLAs, eight LIBERO tasks and five shift settings from four families, we show that ReGround recovers \num{0.91--1.00} of the lost success and is safer than linear calibration, using only a minute of babbling and no gradients, rewards or demonstrations.


\acknowledgments{This work was supported by the Office of Naval Research under Grant N000142412405 and the Army Research Office under Grant W911NF2310363.}


\bibliography{example,extra}  

\clearpage
\appendix

\section{Additional Experiments}
\label{app:extra}
\begin{figure}[h]
  \centering
  \includegraphics[width=\linewidth]{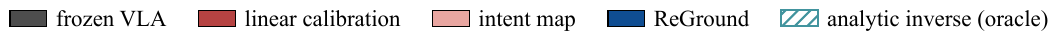}
  \begin{subfigure}[b]{0.49\linewidth}
    \includegraphics[width=\linewidth]{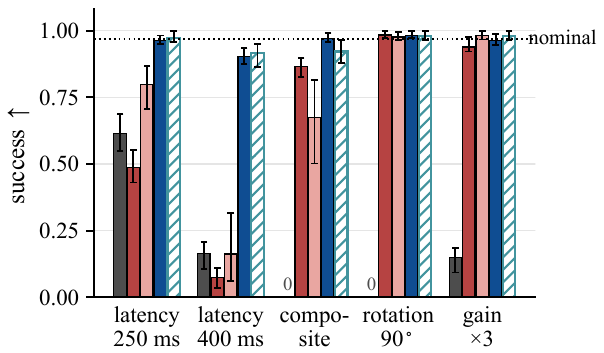}
    \caption{Success on test.}
  \end{subfigure}\hfill
  \begin{subfigure}[b]{0.49\linewidth}
    \includegraphics[width=\linewidth]{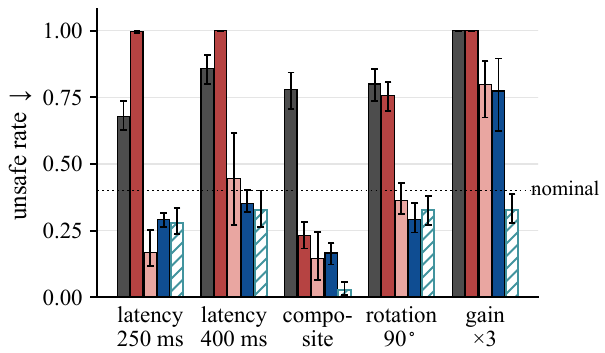}
    \caption{Unsafe rate on test (lower is better).}
  \end{subfigure}\\[4pt]
  \begin{subfigure}[b]{0.49\linewidth}
    \includegraphics[width=\linewidth]{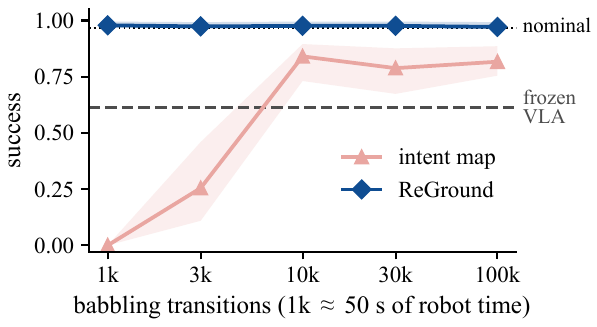}
    \caption{Success vs.\ babbling budget, latency 250\,ms.}
  \end{subfigure}\hfill
  \begin{subfigure}[b]{0.49\linewidth}
    \includegraphics[width=\linewidth]{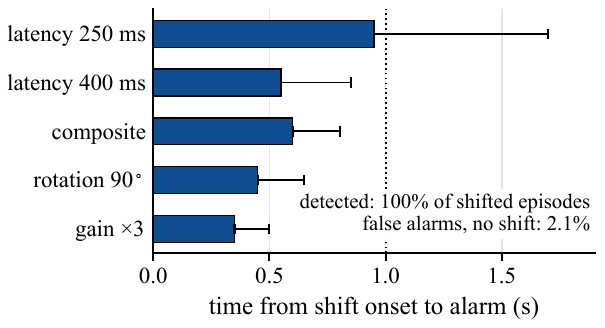}
    \caption{Detection delay of the shift monitor.}
  \end{subfigure}
  \caption{(a,b) Test success and unsafe-episode rate, IQM over 8 tasks at test time. (c) Success versus babbling budget at 250\,ms latency. (d) Detection delay of the shift monitor.}
  \label{fig:effects}
\end{figure}

\begin{table}[h]
  \centering
  \caption{Per-task test success rate of frozen $\pi_{0.5}$ (base) and ReGround (RG). }
  \label{tab:tasks}
  \scriptsize\setlength{\tabcolsep}{2.5pt}
  \resizebox{\linewidth}{!}{
  \begin{tabular}{lcccccccccc}
\toprule
 & \multicolumn{2}{c}{latency 5} & \multicolumn{2}{c}{latency 8} & \multicolumn{2}{c}{composite} & \multicolumn{2}{c}{frame rot.\ 90$^\circ$} & \multicolumn{2}{c}{gain $\times$3}\\
\cmidrule{2-3}\cmidrule{4-5}\cmidrule{6-7}\cmidrule{8-9}\cmidrule{10-11}
task & base & \textbf{RG} & base & \textbf{RG} & base & \textbf{RG} & base & \textbf{RG} & base & \textbf{RG}\\
\midrule
spatial\_0 & 0.23 & 0.99 & 0.00 & 0.91 & 0.03 & 1.00 & 0.00 & 0.99 & 0.09 & 1.00\\
spatial\_2 & 0.54 & 0.95 & 0.00 & 0.83 & 0.00 & 0.98 & 0.00 & 1.00 & 0.03 & 1.00\\
object\_1 & 0.51 & 0.99 & 0.14 & 0.89 & 0.00 & 0.99 & 0.00 & 1.00 & 0.14 & 0.97\\
object\_3 & 0.66 & 0.95 & 0.11 & 0.87 & 0.00 & 0.96 & 0.00 & 0.99 & 0.03 & 0.95\\
goal\_1 & 0.74 & 0.95 & 0.40 & 0.93 & 0.37 & 0.95 & 0.00 & 0.99 & 0.34 & 1.00\\
goal\_2 & 0.74 & 0.94 & 0.43 & 0.95 & 0.00 & 0.92 & 0.00 & 0.95 & 0.34 & 0.90\\
10\_1 & 0.51 & 0.85 & 0.00 & 0.76 & 0.00 & 0.91 & 0.00 & 0.88 & 0.00 & 0.90\\
10\_3 & 0.91 & 0.94 & 0.66 & 0.99 & 0.00 & 0.94 & 0.00 & 0.77 & 0.34 & 0.49\\
\bottomrule
\end{tabular}
  }
\end{table}

\end{document}